%% file: main.tex
\documentclass{article}
\usepackage{amssymb}
\usepackage{amsmath}

\usepackage{amssymb}
\usepackage{graphicx}
\usepackage{booktabs}      
\usepackage{graphicx}       
\usepackage{colortbl}       
\usepackage[table,xcdraw]{xcolor}  
\usepackage{multirow}      
\usepackage{caption}  
\definecolor{dcolor}{RGB}{204,229,255}
\definecolor{cyan1}{RGB}{0,255,255}
\definecolor{codegreen}{rgb}{0.0,0.6,0.0}

\usepackage[final]{corl_2025}  

\title{QuaDreamer: Controllable Panoramic Video Generation for Quadruped Robots}

\author{
Sheng Wu$^{1,*}$, Fei Teng$^{1,*}$, Hao Shi$^{2,3,*}$, Qi Jiang$^{2}$, Kai Luo$^{1}$, Kaiwei Wang$^{2}$, Kailun Yang$^{1,\dag}$\\
$^1$Hunan University \quad $^2$Zhejiang University \quad $^3$Nanyang Technological University\\
}
\makeatletter
\def\thanks#1{\protected@xdef\@thanks{\@thanks
        \protect\footnotetext{#1}}}
\makeatother
\thanks{$^{*}$Equal Contribution. $^{\dag}$Correspondence to kailun.yang@hnu.edu.cn.}

\begin{document}
\maketitle

\vskip-3ex

\begin{figure*}[h]
  \centering
  \includegraphics[width=0.98\textwidth]{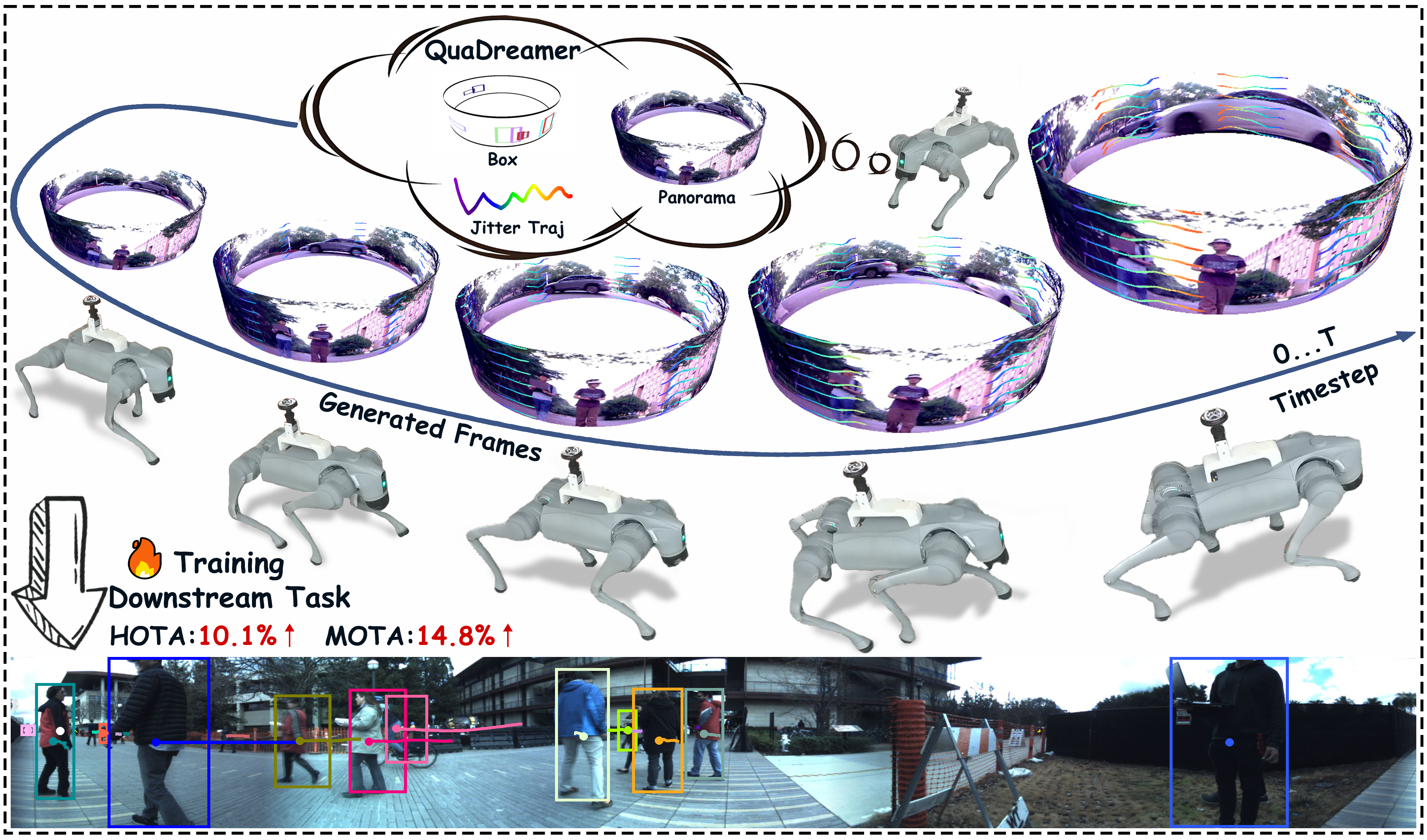}
  \caption{Illustration of the proposed QuaDreamer, the first panoramic video generation framework tailored for quadruped robots. QuaDreamer enables generation with control via box and jitter prompts, serving as a data source to enhance the performance of downstream tasks.}
  \label{fig:Fig_task}
  \vskip -3ex
\end{figure*}

\begin{abstract}
\input{Contents/1_Abst}
\end{abstract}



\section{Introduction}
\input{Contents/2_Into}


\section{Related Work}
\label{sec:Rela}
\input{Contents/3_Rela}
\section{Methodology}
\label{sec:Meth}
\input{Contents/4_Meth}


\section{Experimental Results}
\label{sec:Exps}
\input{Contents/5_Exps}

\section{Conclusion}
\label{sec:Conc}
\input{Contents/6_Conc}
\clearpage

\section{Limitation}
\label{sec:limit}
\input{Contents/7_limit}

\acknowledgments{This work was supported in part by the National Natural Science Foundation of China (Grant No. 62473139), in part by the Hunan Provincial Research and Development Project (Grant No. 2025QK3019), and in part by the Open Research Project of the State Key Laboratory of Industrial Control Technology, China (Grant No. ICT2025B20).}


\bibliography{Reference}  

\section{Experimental Details}

\input{Contents/1_Exp}
\section{PTrack Evaluation Method}
\input{Contents/2_PTrack}
\section{Controllability Analysis}
\input{Contents/3_Control}

\section{Visualization Results}
\input{Contents/5_Visual}

\section{Research Significance}
\input{Contents/6_Significance}
\section{Future Work}
\input{Contents/7_Futurework}
\end{document}

%% file: Contents/1_Abst.tex
Panoramic cameras, capturing comprehensive 360-degree environmental data, are suitable for quadruped robots in surrounding perception and interaction with complex environments. However, the scarcity of high-quality panoramic training data — caused by inherent kinematic constraints and complex sensor calibration challenges — fundamentally limits the development of robust perception systems tailored to these embodied platforms. To address this issue, we propose QuaDreamer—the first panoramic data generation engine specifically designed for quadruped robots. QuaDreamer focuses on mimicking the motion paradigm of quadruped robots to generate highly controllable, realistic panoramic videos, providing a data source for downstream tasks. Specifically, to effectively capture the unique vertical vibration characteristics exhibited during quadruped locomotion, we introduce Vertical Jitter Encoding (VJE). VJE extracts controllable vertical signals through frequency-domain feature filtering and provides high-quality prompts. To facilitate high-quality panoramic video generation under jitter signal control, we propose a Scene-Object Controller (SOC) that effectively manages object motion and boosts background jitter control through the attention mechanism. To address panoramic distortions in wide-FoV video generation, we propose the Panoramic Enhancer (PE) – a dual-stream architecture that synergizes frequency-texture refinement for local detail enhancement with spatial-structure correction for global geometric consistency. We further demonstrate that the generated video sequences can serve as training data for the quadruped robot's panoramic visual perception model, enhancing the performance of multi-object tracking in 360-degree scenes. The source code and model weights will be publicly available at \url{https://github.com/losehu/QuaDreamer}.

%% file: Contents/2_Into.tex
Quadruped robots have become a crucial component of advanced embodied intelligence technologies, demonstrating significant potential across a variety of application scenarios such as inspection\cite{parkinson2023assessment,torres2024investigating}, search and rescue~\cite{patel2022lidar}, and security~\cite{saraf2021terrain}. An increasing number of researchers are focusing on enhancing robotic agents’ understanding of their environments~\cite{xu2023understanding,mudalige2022dogtouch,li2023seeing,ouyang2024long}. 
With the advantages of a higher Field-of-View (FoV) relative to device size ratio~\cite{yang2021context,yang2020ds,tateno2018distortion,shen2022panoformer,ai2025survey} and better alignment with LiDARs, panoramic cameras with a 360-degree FoV offer a convenient solution for quadruped robots to achieve a comprehensive understanding of the surrounding scenes.
However, an embodied agent’s ability to understand its environment is fundamentally a data-driven process, heavily reliant on training with accurately annotated datasets~\cite{gao2023magicdrive}. 
Large-scale, diverse, and photorealistic data is essential for achieving robust perception in dynamic environments.
Currently, panoramic datasets captured from the perspective of quadruped robots remain scarce due to multiple challenges, including the limited endurance of quadruped platforms~\cite{miller2020mine} and the instability of the ego-agent~\cite{saraf2021terrain}. 
These challenges introduce the issue of homogeneous scene distribution in existing datasets~\cite{OminiTrack}, along with the technical difficulties of image stitching and time synchronization~\cite{wei2024onebev}, making the data collection process highly labor-intensive. 
To enable mass production of high-quality panoramic data for scene understanding of embodied agents, we propose \textbf{QuaDreamer}—the first panoramic motion generation engine (world model) specifically designed for quadruped robots, as shown in Fig.~\ref{fig:Fig_task}. 
By providing a single panoramic image, object motion trajectories, and the vibration signals generated during the robot's movement, high-quality panoramic videos are generated to recreate the motion and vibration characteristics of the quadruped robot. QuaDreamer consists of three main components: the Vertical Jitter Encoding (VJE), Scene-Object Controller (SOC), and Panoramic Enhancer (PE).

Specifically, VJE employs a high-pass filter to decouple the low-frequency object-relative trajectories from the high-frequency vertical jitter. The extracted jitter signal represents spatial-domain positional offsets of the quadruped robot caused by locomotion dynamics. Subsequently, these spatial offsets are projected into the feature domain via a camera encoder, enabling generation models to explicitly learn and leverage vibration patterns in both geometric and latent spaces. In latent spaces, control signals are integrated using an attention mechanism, while in the geometric space, a position-aware diffusion process is constructed: by using the SOC to merge the object bounding box with the features encoded by the camera, and seamlessly integrating this into the pre-trained video diffusion model, jitter control is enhanced, enabling precise object position control. Meanwhile, panoramic imagery substantially expands the FoV but simultaneously introduces considerable challenges to maintaining global consistency during the diffusion process. To address this challenge, we propose the Panoramic Enhancer, a dual-stream module that simultaneously processes frequency-domain features and spatial-domain structures. This module includes a Fourier CNN Texture Restorer (FTR)~\cite{dong2022incremental}, which employs frequency-domain convolutions with global receptive fields to mitigate resolution-sensitive artifacts while preserving high-frequency details, and a State Space Model (SSM)~\cite{dao2024transformers}, which enforces structural continuity across distorted regions through selective state transitions. The bidirectional interaction between these components allows for effective resolution of both local distortions and global inconsistencies.
To verify the effectiveness of our proposed method, we evaluate the model in terms of controllability and video quality, ensuring a comprehensive assessment of the framework. Specifically, the LPIPS~\cite{zhang2018unreasonable} score decreases by $3.68\%$ and the SSIM~\cite{wang2004image} score increases by $3.46\%$ compared to our baseline. On the evaluation metric PTrack designed for assessing controllability, QuaDreamer improves by $43.86\%$ over the baseline, demonstrating a significant enhancement in synthesized video control. Furthermore, our architecture without control modules outperforms SVD in terms of video quality.
We also evaluate the synthesized videos on downstream tasks, demonstrating that the generated videos serve effectively as data augmentation, benefiting multi-object tracking tasks in complex, unconstrained surroundings.

In summary, the contributions of this work are as follows:
\begin{itemize}
\item We introduce QuaDreamer, the first panoramic data generation engine specifically designed for quadruped robots, capable of generating controllable panoramic videos with natural vibrations from the perspective of a quadruped robot.
\item An in-depth analysis of control signals is presented, with Vertical Jitter Encoding (VJE) put forward to extract jitter signals. The Scene-Object Controller (SOC) is developed to effectively regulate video vibrations and object movements. 
Additionally, we propose a dual-stream design that integrates frequency-domain features and spatial-domain structures — Panoramic Enhancer (PE), aimed at enhancing image quality.

\item Through extensive experiments, we demonstrate the effectiveness of QuaDreamer in achieving precise controllability. Additionally, our results indicate that the synthetic data significantly enhances performance in downstream multi-object tracking tasks for quadruped robots, with HOTA improving by $10.14\%$ and MOTA improving by $14.75\%$.
\end{itemize}

%

%% file: Contents/3_Rela.tex
\textbf{Perception for Quadruped Robots.}
Quadruped robots are increasingly vital in embodied AI for tasks like disaster response~\cite{oh2024trip,patel2022lidar,lee2025trg}, assistive navigation for the visually impaired~\cite{hwang2024lessons}, and industrial inspection~\cite{parkinson2023assessment,lee2024safety,torres2024investigating}. Enhancing their perception~\cite{mudalige2022dogtouch,bellegarda2024visual,elnoor2024amco} is critical, with advances in visual navigation~\cite{dao2024transformers}, vision-language models~\cite{ding2024quar}, and object tracking~\cite{OminiTrack}. Panoramic perception~\cite{gao2022review,shi2023panoflow,zhang2024behind} is particularly beneficial for quadrupeds, offering comprehensive situational awareness essential for navigating complex terrains and managing dynamic body movements inherent to their locomotion. Complementing perception, accurate state estimation is fundamental for stable control and autonomous navigation~\cite{nistico2025muse,menner2024simultaneous}. Methods often achieve this by fusing data from various sensors, such as cameras (including panoramic~\cite{wang2022lf,wang2023lf,wang2024lf}), LiDAR, IMUs, and leg odometry, using techniques like factor graph optimization~\cite{wisth2019robust,wisth2022vilens} or specialized algorithms~\cite{kim2021legged}.
Despite progress, the complex motion patterns of quadrupeds present significant hurdles for data collection and annotation, especially for multi-sensor or panoramic systems. Furthermore, inherent vibrations can degrade camera data quality, particularly impacting wide-fov panoramic sensors. To overcome these data bottlenecks, our work focuses on efficiently and affordably generating controllable, high-quality panoramic videos from existing data. This approach preserves quadruped-specific motion consistency while enabling diverse styles, providing a valuable resource for advancing perception research for these agile robots.

\textbf{Diffusion Models for Conditional Generation.} The diffusion model~\cite{ho2020denoising,song2020score,zheng2023non} was originally designed for image generation by learning the gradual denoising process from a Gaussian noise distribution to an image distribution. This paradigm achieves great success in high-quality video generation~\cite{ho2022video,blattmann2023stable,ho2022video}, such as SVD~\cite{blattmann2023stable}, AnimateDiff~\cite{guo2023animatediff}, and VideoCrafter~\cite{chen2024videocrafter2}. 
Based on that, several works have introduced control signals and diverse control mechanisms to generate diverse controllable videos, such as camera pose-driven video generation~\cite{he2024cameractrl,wang2024motionctrl,zheng2024cami2v,xu2024camco}, trajectory-based video generation~\cite{wang2024framer,wu2024draganything,wang2024objctrl,yin2023dragnuwa}, and layout-to-image generation~\cite{li2023gligen,cheng2023layoutdiffuse,gao2023magicdrive,li2024trackdiffusion}. Furthermore, works such as GeoDiffusion~\cite{chen2023geodiffusion}, TrackDiffusion~\cite{li2024trackdiffusion}, and MagicDrive~\cite{gao2023magicdrive} have shown that generated data can serve as datasets to boost downstream tasks. However, these works focus on global style and viewpoint transformations but fail to capture the subtle spatial jitters typical of quadruped robots. Therefore, this paper explores the potential of Image-to-Video (I2V) diffusion models in generating panoramic videos from the perspective of a quadruped robot, and, based on this, to advance the development of downstream perception models for quadruped robots.
%

%% file: Contents/4_Meth.tex
\begin{figure*}[ht]
  \centering
  \includegraphics[width=1\textwidth]{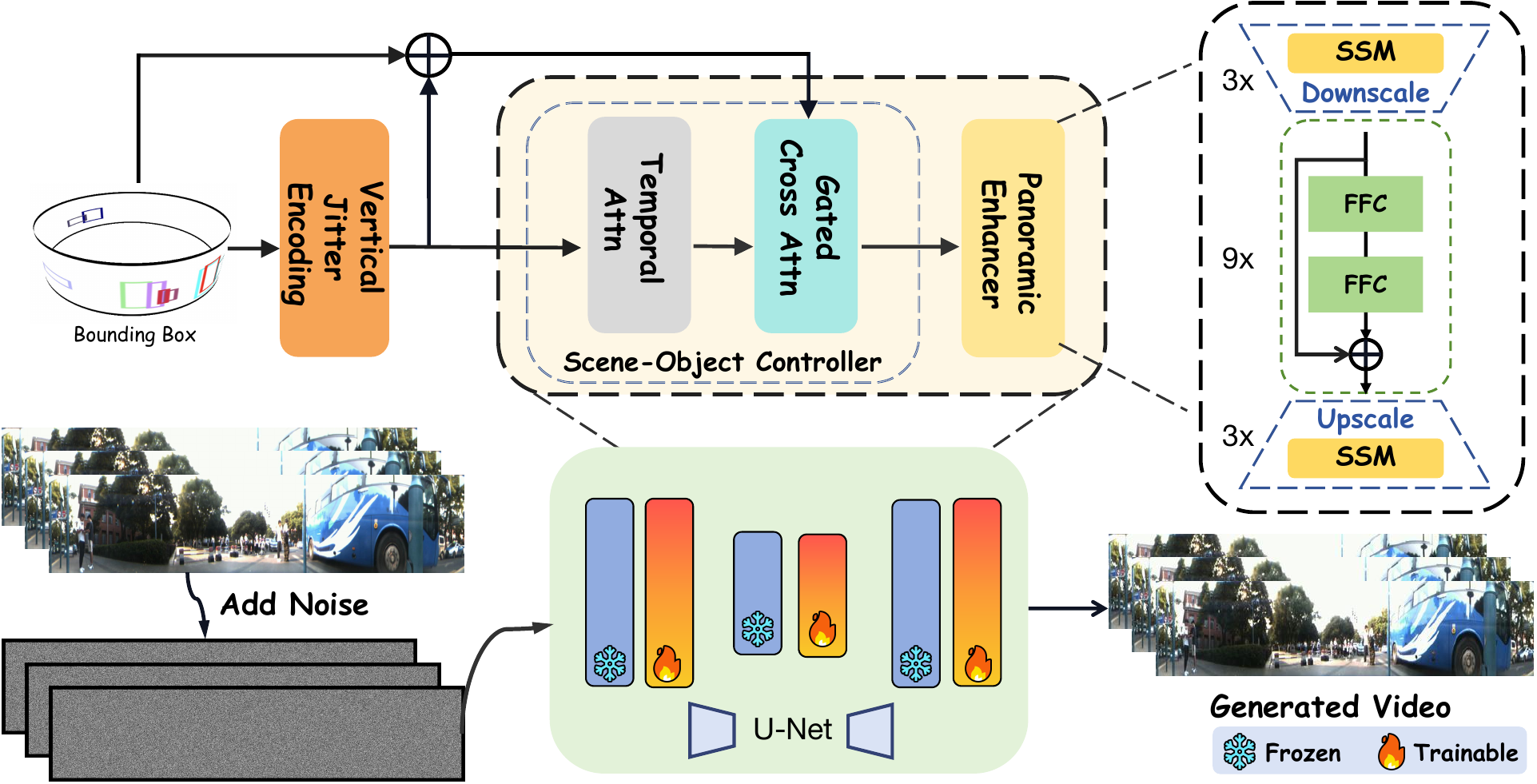}
  \vskip-1ex
  \caption{\textbf{The overall framework of our QuaDreamer.} Vertical Jitter Encoding extracts jitter signals from bounding boxes and combines them with box information to accurately model motion patterns. To further enhance realism, we incorporate a Scene-Object Controller and a Panoramic Enhancer, which jointly manage object dynamics and refine the representation of panoramic motion.}
  \label{fig:Fig2}
  \vskip -2ex
\end{figure*}

To generate controllable panoramic videos from the quadruped robot's perspective under intense motion jitter, we propose the QuaDreamer architecture. QuaDreamer integrates control signals into the Latent Diffusion Model~\cite{rombach2022high}. VJE efficiently extracts jitter signals and projects them into the feature domain. The SOC seamlessly integrates jitter and geometric features into U-Net~\cite{ronneberger2015u} through an attention mechanism. 
To enhance image quality and adapt to distortion, a dual-stream structure, PE, is designed to refine image quality in the frequency domain and accommodate panoramic distortion in the spatial domain. 
The architecture of QuaDreamer is shown in Figure~\ref{fig:Fig2}.

\textbf{Preliminary.}
Latent Diffusion Models~\cite{rombach2022high} combine the advantages of autoencoders and diffusion models by performing the diffusion process in the latent space, significantly reducing computational cost while maintaining high-quality image generation.
During the diffusion process, noise is progressively added to the latent representation \( z \) until it transforms into standard Gaussian noise at step \( T \). Specifically, starting from the initial latent representation \( z_0 \sim q(z_0) \), the latent representation \( z_t \) is iteratively updated as follows:
\begin{equation}
q(z_t | z_{t-1}) = \mathcal{N}(z_t; \sqrt{1 - \beta_t} z_{t-1}, \beta_t I), \quad t = 1, \dots, T,
\end{equation}
where \( \beta_t \) regulates the strength of the noise added at each step during the diffusion process.
A U-Net is trained to predict the denoising process, and the MSE loss is used to optimize the network:
\begin{equation}
\mathcal{L}(\theta) = \mathbb{E}_{t \sim u(1, T), \epsilon_t \sim \mathcal{N}(0, I)} \left[ \| \epsilon_t - \epsilon_\theta(z_t; t, y) \|^2 \right],
\end{equation}
where \( y \) represents the conditional information. By minimizing the loss value, the network learns to reconstruct latent representations that closely match the original data.

\textbf{Vertical Jitter Encoding.} 
Since the data collector is always present in the panoramic images, we decompose the vertical coordinates of the collector's bounding box \( y_{\text{human}}(t) \) into low-frequency and high-frequency components through spectral analysis.  
The low-frequency component represents the relative displacement between the collector and the quadruped robot, exhibiting stable, slow-changing characteristics. The high-frequency component reflects the vertical jitter induced by the gait characteristics of the quadruped robot, manifested as oscillatory fluctuations in the vertical direction, as shown in Figure~\ref{fig:Fig3}.  
To extract the high-frequency part, we used a first-order Butterworth high-pass filter~\cite{butterworth1930theory}. The frequency response \( H(f) \) of the filter is given by the following formula:
\begin{figure*}[ht]
  \centering
  \includegraphics[width=1\textwidth]{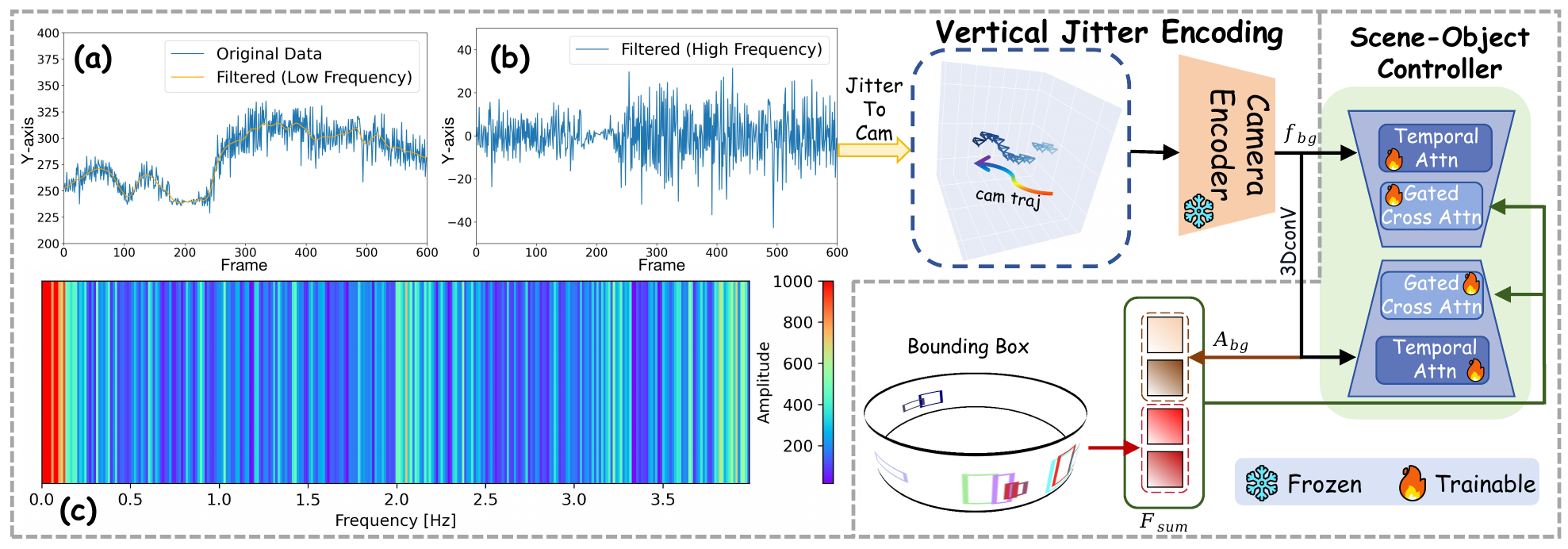}
  \vskip-1ex
  \caption{\textbf{Illustration of the control components, including VJE and SOC.} (a) shows the original y-axis pixel coordinate data and its low-frequency component; (b) displays the filtered high-frequency jitter data; (c) illustrates the frequency spectrum of the original data.}
  \label{fig:Fig3}
  \vskip -3ex
\end{figure*}
\begin{equation}
H(f) = \frac{(f / f_c)^{2n}}{1 + (f / f_c)^{2n}},
\end{equation}
where \( f \) is the frequency, \( f_c \) is the cutoff frequency, and \( n \) is the order of the filter. The cutoff frequency \( f_c \) was set to 0.3 Hz, and the filter order \( n \) was set to 1.0.
By applying this high-pass filter, we obtained the filtered signal \( y_{\text{w}}(t) \), which represents the extracted vertical jitter data of the quadruped robot, expressed as:
\begin{equation}
y_{w}(t) = \mathcal{F}^{-1}\left[\mathcal{F}(y_{\text{human}}(t)) \cdot H(f)\right],
\end{equation}
where \( \mathcal{F} \) and \( \mathcal{F}^{-1} \) represent the Fourier transform and its inverse, respectively. 

We first convert the vertical jitter data into the world coordinate system, and the world coordinates for each frame are represented as \( P_w^i = (x_w^i, y_w^i, z_w^i) \), where \( x_w^i \) is set to half the image width, \( z_w^i \) is fixed at 1 to represent the global depth information of the image, and \( y_w^i = y_{\text{w}}(i) \) represents the vertical jitter data in the world coordinate system. Next, using the camera intrinsic matrix \(K = [[f_x, 0, c_x],[ 0, f_y, c_y],[ 0, 0, 1]]\), the world coordinates \( P_w^i  \) can be mapped into the camera coordinate system \( C^i = (x^i_c, y^i_c, z^i_c) \). Following previous works~\cite{wang2024shape,wang2024videocomposer}, the matrix \( K \) can be roughly estimated based on the spatial dimensions of the generated video. Using the formula \( C^i = K \cdot P_w^i \), we convert the world coordinates \( P_w^i\) into the camera coordinates \( C^i \).
Compared to directly using camera coordinates, Plücker embeddings~\cite{sitzmann2021light} enable more precise control over visual details by effectively representing camera pose. This avoids the imprecision of raw parameters while allowing efficient per-pixel manipulation in image space, thereby enhancing visual consistency and accuracy.
To leverage this, we first convert the camera pose $C^i$ into its Plücker embedding $P_{u}^i$. Then, the camera encoder processes this representation to compute per-frame background features, yielding $f^{i}_{bg} = \text{PoseEncoder}(P_{u}^i).$

\textbf{Scene-Object Controller.} 
The core idea of SOC is to decompose the scene into two interrelated components: the background motion field and the object motion field. Specifically, the background motion field is modeled by embedding the background features \( f_{\text{bg}}^i \) into a temporal attention module to capture the temporal dependencies of the camera trajectory. These features are combined with the image latent features through pixel-level addition, enhancing the control over background jitter. On the other hand, the background motion field is also processed using 3D convolutions to extract spatiotemporal features \( A_{\text{bg}}^i = \text{Conv3D}(f_{\text{bg}}^i) \), modeling camera jitter and global motion patterns.

The operation aggregates features across both time and space, enabling the network to learn the time-varying spatial jitter characteristics. Meanwhile, the object motion field is encoded using Fourier Embedding, which maps the bounding box coordinates into a high-dimensional frequency space. This representation enables it to seamlessly integrate with spatiotemporal features in subsequent operations, as expressed by:
\begin{equation}
\Gamma(b_t) =  \bigoplus_{k=1}^{K } \left[ \sin(\omega_k b_t), \cos(\omega_k b_t) \right],
\end{equation}
where \( \omega_k = \tau^{\frac{k}{K}} \) forms geometric frequency bands (\(\tau = 100\)), allowing us to capture the multi-scale nature of object motion. Additionally, by incorporating the binary switching mechanism of the visibility mask \( m_t \in \{0, 1\} \), the model is able to learn differentiated motion representations of the object in both visible and occluded states:
\begin{equation}
B_t = m_t \cdot \Gamma(b_t) + (1 - m_t) \cdot \varphi_{\text{null}},
\end{equation}

Here, \( \varphi_{\text{null}} \) represents a learnable zero embedding, used to fill in the missing detection frame information. To integrate object motion with background dynamics, the object motion field \( \mathbf B = \{ B_0, B_1, \dots, B_t \} \) is fused with the background features \(\mathbf {A_{\text{bg}}} = \{A_{\text{bg}}^0 , A_{\text{bg}}^1, \dots, A_{\text{bg}}^i \} \), generating the final feature representation:
\begin{equation}
F_{\text{sum}} = [\text{MLP}(\text{Fourier}(\mathbf B)), \text{MLP}(\mathbf{A_{\text{bg}})}].
\end{equation}
The fused representation \(F_{\text{sum}}\) is subsequently passed through a gated self-attention mechanism~\cite {li2023gligen} to integrate it with the input visual features.

\textbf{Panoramic Enhancer.} 
The proposed Panoramic Enhancer (PE) is based on an encoder-decoder architecture, which injects the State Space Model (SSM)~\cite{dao2024transformers} symmetrically at specific layers: before the first downsampling layer and after the final downsampling layer. This design performs frequency domain optimization at the intermediate layers, collaboratively addressing the issues of geometric distortion and detail degradation in wide-FoV image generation~\cite{OminiTrack}. As illustrated in Figure~\ref{fig:Fig2}, the input image first passes through an SSM module at the initial encoder layer, where multi-directional scanning mechanisms model long-range geometric dependencies to suppress edge distortions caused by panoramic unwrapping. For the feature tensor \( D \in \mathbb{R}^{B \times C \times H \times W} \) from the encoder downsampling layer, the SSM operation is defined as follows:
\begin{equation}
D^*[b, c, x, y] = \frac{1}{N} \sum_{d \in \text{scan}} F_{S6}(S_d(D[b, c, x, y])),
\end{equation}
where \( N \) represents the number of scans, \( S_d \) represents the scanning function, and \( F_{S_6} \) is the transformation function for the S6 block~\cite{dao2024transformers}.
Following three successive downsampling stages, the intermediate layers further process the output features. To enhance frequency-domain representations, we employ residual blocks with Fast Fourier Convolution (FFC)~\cite{suvorov2022resolution}:
\begin{equation}
\label{eq:ffc}
X_{out} = X + FFC(FFC(X_l, X_g)),
\end{equation}
where $X = (X_l, X_g)$ is the input divided into local and global components at a $25\%$ and $75\%$ ratio~\cite{suvorov2022resolution}. This design enables effective recovery of periodic structures while avoiding grid artifacts~\cite{dong2022incremental}.
The decoder progressively reconstructs high-resolution images through a three-stage upsampling process, with a secondary SSM injection applied after the final upsampling layer to enhance geometric continuity. This hierarchical architecture enables phased collaboration between spatial-domain correction and frequency-domain refinement: geometric constraints propagate bidirectionally through SSM-enhanced encoder-decoder layers, while FFC-driven intermediate optimization focuses on detail enhancement.

\begin{figure*}[!t]
  \centering
  \includegraphics[width=1.0\textwidth]{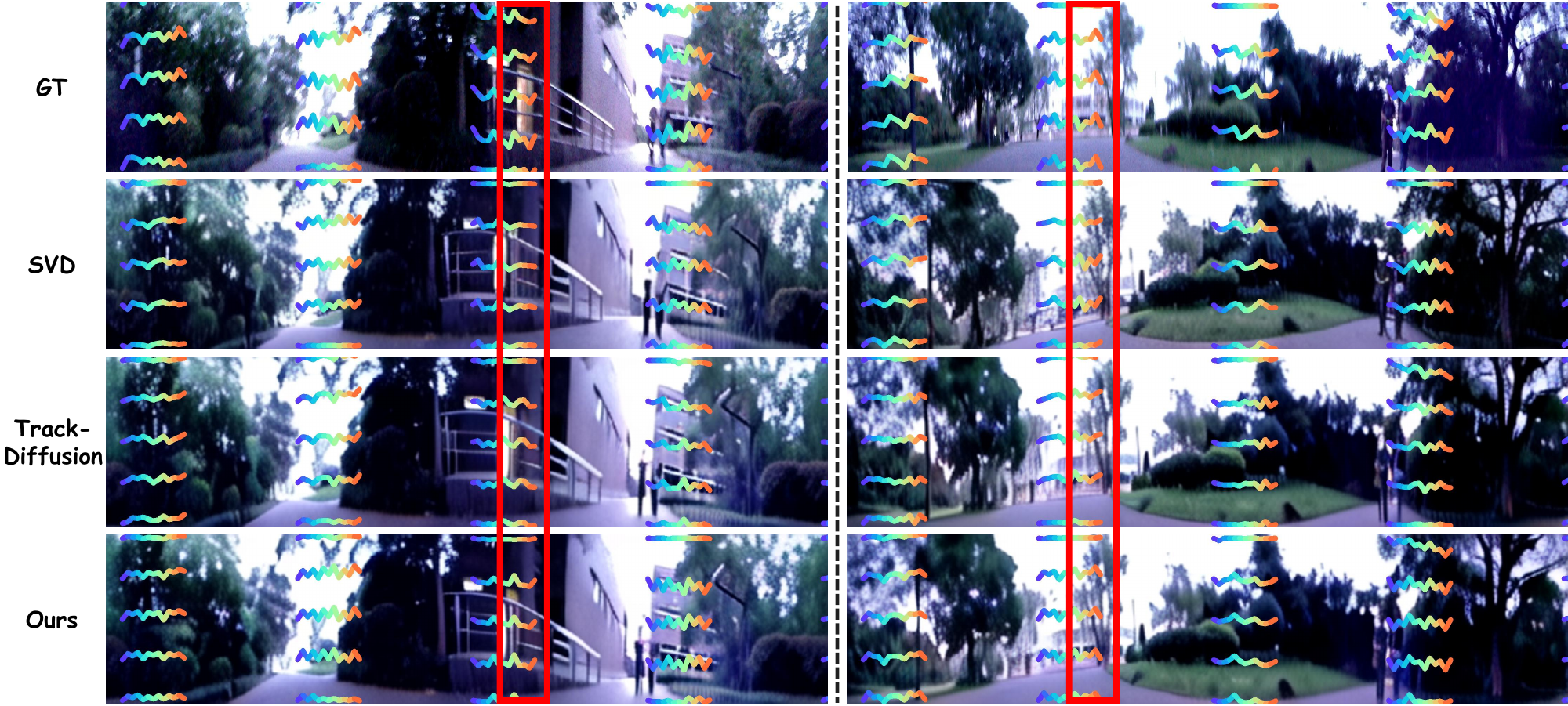}
  \vskip-1ex
  \caption{
  \textbf{Visualization results generated on the QuadTrack dataset~\cite{OminiTrack}.}
  The rainbow-colored trajectory is the CoTracker~\cite{karaev2024cotracker3} jittering trajectory.
  The trajectory in the \textcolor{red}{\boxed{\textcolor{black}{red\ boxes}}} clearly demonstrates that our jitter control is more similar to the ground truth.}
  \label{fig:Fig4}
  \vskip -3ex
\end{figure*}

%% file: Contents/5_Exps.tex
\textbf{Dataset and Baselines.}
The experiments are conducted on the QuadTrack dataset~\cite{OminiTrack}, which contains $19,200$ panoramic frames captured from the quadruped robot's perspective, along with $189,876$ high-quality manually annotated object bounding boxes. 
The training set comprises $702$ video sequences, each with $8$ frames, and an additional $3,600$ frames are reserved separately for validating downstream tasks. The test set includes $378$ sequences (each with $14$ frames), specifically used for evaluating generation quality.
We adopt TrackDiffusion~\cite{li2024trackdiffusion} as the baseline method, where the model inputs consist of panoramic images and object bounding boxes. Additional implementation details are provided in the supplementary material.

\textbf{Evaluation Metrics.}
We employed FVD~\cite{unterthiner2018towards}, LPIPS~\cite{zhang2018unreasonable}, PSNR, and SSIM~\cite{wang2004image} to assess the video quality. 
To evaluate the controllability, \textit{i.e.}, the correspondence between the input bounding boxes and the generated instances, we used tracking metrics such as HOTA~\cite{luiten2021hota} and MOTA~\cite{bernardin2008evaluating}. 
These metrics were applied using the OmniTrack model~\cite{OminiTrack} to track objects in the generated video, and then compared with the ground truth tracklets.
Additionally, we utilized CoTracker~\cite{karaev2024cotracker3} to measure point-tracking consistency for assessing video jitter control performance, denoted as PTrack.
Following the conventions of TrackDiffusion~\cite{li2024trackdiffusion} and MagicDrive~\cite{gao2023magicdrive}, we augmented the training set for downstream tasks with the generated images, validating their contribution to improving task performance.

\begin{table}[h]
\vskip-1ex
\centering
\renewcommand{\arraystretch}{1.2}
\setlength{\tabcolsep}{1.05mm}
\begin{tabular}{@{}l|>{\centering\arraybackslash}p{1.3cm}|cccc|ccc@{}}
\toprule
\textbf{Method} & \textbf{Control} & FVD ↓ & LPIPS ↓ & PSNR↑ & SSIM↑ & HOTA ↑ & MOTA ↑ & PTrack↓ \\
\midrule
SVD~\cite{blattmann2023stable}  &\(\times\)  & 831.31 &  0.2711 &  14.50  & 0.3874  & 9.6113& -22.927 & \textbf{24.1699} \\

\textbf{QuaDreamer*} &\(\times\)  &  \textbf{830.92} &\textbf{0.2669} &  \textbf{14.51}  & \textbf{0.3893}  &\textbf{10.332}& \textbf{-20.395} & 25.5571 \\

\midrule

TrackDiffusion~\cite{li2024trackdiffusion}  &  \small{box}  & \textbf{887.31} &  0.2714 &  14.27  & 0.3815  & 9.6528& -21.451 &25.2496 \\
\textbf{QuaDreamer (Ours)}& \small{box\&jitter}& 895.70 &  \textbf{0.2614} &  \textbf{14.51}  & \textbf{0.3947}  &  \textbf{9.6542}& \textbf{-21.285} &\textbf{14.1744} \\ 
\bottomrule
\end{tabular}
\vspace{2mm}
\caption{\textbf{Comparison of generation fidelity on the QuadTrack dataset}~\cite{OminiTrack}. 
TrackDiffusion~\cite{li2024trackdiffusion} is our baseline.
\textbf{QuaDreamer*} is a model that does not integrate any control modules.}
\label{tab:method_comparison}
\vskip-3ex
\end{table}

\subsection{Main Results} The Tab.~\ref{tab:method_comparison} presents the video quality and controllability test results of our proposed QuaDreamer and other baseline models on the QuadTrack dataset. The visualization results are shown in Figure~\ref{fig:Fig4}. Through comparative experiments with the uncontrolled SVD~\cite{blattmann2023stable} method of the same type, the effectiveness of the Panoramic Enhancer in improving video quality is validated. Compared to baseline methods, QuaDreamer demonstrates superior overall performance. Specifically, in terms of image quality, LPIPS decreased by $3.68\%$, SSIM increased by $3.46\%$, and PSNR improved by $1.68\%$. 
Meanwhile, our method is the first to achieve controllable generation based on jitter. Although FVD increased slightly, this is primarily due to the rise in control dimensions. The significant reduction in the PTrack metric (computed by tracking frame-wise jitter with CoTracker~\cite{karaev2024cotracker3}) effectively validates the controllability of high-frequency jitter in the generated quadruped robot videos.

\subsection{Ablation Study}

\textbf{Ablation of QuaDreamer Framework.}
In Tab.~\ref{tab:qua_ablation}, to verify the effectiveness of our designed modules, we conducted an ablation study on two key modules. Setting (a) represents the baseline method, illustrating the model's initial performance. In setting (b), SOC was used, and improvements in MOTA (increased by $6.74\%$) and PTrack (decreased by $36.76\%$) performance were observed, indicating an enhancement in control effectiveness. In setting (c), the introduction of PE led to a significant improvement in FVD, with a decrease of $6.36\%$. Finally, setting (d) represents our final model, which integrates both modules, thereby enhancing both controllability and video quality.

\begin{table}[h]
\vskip-1ex
\centering
\renewcommand{\arraystretch}{1.2}
\setlength{\tabcolsep}{1.5mm}  
\begin{tabular}{@{}cw{c}{1cm}w{c}{1cm}|cccc|ccc@{}}
\toprule

Setting & SOC & PE  &  FVD ↓ & LPIPS ↓ &PSNR↑&  SSIM↑  & HOTA ↑ & MOTA ↑& PTrack↓ \\
\midrule

(a) &&  & 887.31 &  0.2714 &  14.27  & 0.3815  & 9.6528& -21.451 &25.2496 \\
(b) & \checkmark &  &  896.07 & 0.2620  & 14.51 & 0.3950 &9.3924 & \textbf{-20.006} &15.9676 \\
(c) &  & \checkmark &  \textbf{830.92}& 0.2669& 14.51 &0.3893& \textbf{10.332} & -20.395 &25.5571 \\
\rowcolor{gray!20} (d) & \checkmark & \checkmark & 895.70 &  \textbf{0.2614} &  \textbf{14.51}  & \textbf{0.3947}  &  9.6542& -21.285 &\textbf{14.1744}  \\
\bottomrule
\end{tabular}
\vspace{2mm}
\caption{Ablation of Scene-Object Controller and Panoramic Enhancer. (a) indicates the baseline.} 
\label{tab:qua_ablation}
\vskip-2ex
\end{table}
\textbf{Analysis of Benefits for Downstream Task.}
In this section, we analyze the advantages of using frames generated by QuaDreamer for data augmentation to train a multi-object tracker. 
Experiments are conducted on OmniTrack~\cite{OminiTrack}, an omnidirectional multi-object tracking task specifically for quadruped robots navigating in unconstrained environments. 
In OmniTrack, the model is designed to handle the distortions of panoramic images while maintaining multi-object tracking performance and robustness, even when the robot dog undergoes uneven motion.
We incorporate these generated frames into the training process to explore how generated data contributes to this challenging downstream task. 
Using single-frame images and object bounding box annotations from the QuadTrack test set as input, and referring to MagicDrive~\cite{gao2023magicdrive} and TrackDiffusion~\cite{li2024trackdiffusion}, we generate additional frames for augmented samples and combine them with real images to enrich the training dataset.
Subsequently, the model is trained using the default settings of OmniTrack and evaluated on the validation set. 
The results in Tab.~\ref{tab:tracking_metrics} demonstrate the effectiveness of QuadTrack in enhancing the quadruped robot multi-object tracking model. 
When comparing the model trained with only real data to the model augmented with frames generated by QuaDreamer, a noticeable gain in performance is observed. Specifically, QuaDreamer improved HOTA by $\textbf{1.437}$ and MOTA by $\textbf{10.095}$, indicating a significant increase in tracking accuracy. 
In contrast, the MOTA performance of other models showed a slight decline. 
This indicates that the data generated by QuaDreamer can effectively support downstream tasks for the quadruped robot. 
Its strengthened performance stems from its ability to more realistically simulate jitter and provide higher-quality images, thereby enhancing the data's authenticity and diversity, which helps downstream task models better adapt to complex real-world environments.

\begin{table}[h]
\vskip-1ex
\centering
\renewcommand{\arraystretch}{1.2}
\setlength{\tabcolsep}{1mm}
\begin{tabular}{@{}l|cccc@{}}
\toprule
\textbf{Method} & \textbf{HOTA ↑} & \textbf{MOTA ↑} &\textbf{DetA ↑} &\textbf{AssA ↑} \\
\midrule

Real Only     & 14.168      &   -68.427  & 12.871&	16.052  \\
\midrule

SVD~\cite{blattmann2023stable}           &    14.539\textcolor{codegreen}{(+2.62\%)}     & -72.153\textcolor{red}{(-5.45\%)}  &12.361\textcolor{red}{(-3.96\%)} &	17.666\textcolor{codegreen}{(+10.1\%)}       \\
TrackDiffusion~\cite{li2024trackdiffusion}   &     14.793\textcolor{codegreen}{(+4.41\%)}   &   -72.102\textcolor{red}{(-5.37\%)}  &12.758\textcolor{red}{(-0.88\%)}&	17.596\textcolor{codegreen}{(+9.62\%)}      \\
\rowcolor{gray!20} \textbf{QuaDreamer (Ours)}       &  \textbf{15.605\textcolor{codegreen}{(+10.1\%)}}   &    \textbf{-58.332\textcolor{codegreen}{(+14.8\%)} }  & \textbf{13.595\textcolor{codegreen}{(+5.63\%)}} 	& \textbf{18.384\textcolor{codegreen}{(+14.5\%)}}     \\
\bottomrule
\end{tabular}
\vspace{2mm}
\caption{\textbf{The improvement of downstream task performance by different models.} {``Real Only''} means that OmniTrack is trained without using any data augmentation.}
\label{tab:tracking_metrics}
\vskip-3ex
\end{table}

%% file: Contents/6_Conc.tex
We propose QuaDreamer, the first controllable panoramic data generation engine designed explicitly for quadruped robots. 
QuaDreamer explores a new data extraction method and is capable of producing highly controllable and realistic panoramic videos for perception tasks in complex environments.

%% file: Contents/7_limit.tex
This paper focuses on the generation conditions under the small high-frequency jitter of quadruped robots, but lacks control generation for the movement of other degrees of freedom. In the future, we will explore other control methods for generation models. In addition, leveraging large language models to enable generation under coarsely annotated conditions also presents a valuable and underexplored avenue.

%

%% file: Contents/1_Exp.tex
\textbf{Details.} 
We conducted the full training on a single NVIDIA A6000 48G GPU for $320$ epochs, 
totaling $200,000$ training steps, with the entire process taking $78$ hours. 
The experiment utilized the accelerate library to streamline the training workflow, 
employing a batch size of $2$ and gradient accumulation steps of $1$ to optimize memory utilization. 
We set an initial learning rate of $3e-5$ with a linear warm-up over the first $750$ steps to stabilize early-stage convergence. 
Additionally, mixed-precision training (FP16) was enabled, significantly improving training speed and reducing memory consumption while maintaining model accuracy. 
During generation, frames are sampled using the DPM-Solver~\cite{lu2022dpm} scheduler for $30$ steps. 
Other comparative models were trained and inferred using the same settings to ensure a fair evaluation.
We use the pre-trained Stable Video Diffusion~\cite{blattmann2023stable} as the base generative model, 
combined with the frozen-parameter CameraCtrl camera encoder~\cite{he2024cameractrl} to encode jitter, 
a setup that significantly reduces the number of parameters that need to be trained.

\textbf{Datasets and Evaluation Methods.}
Given that QuadTrack~\cite{OminiTrack} is currently the only existing quadruped robots perspective panoramic dataset, we conduct training and controllability evaluation on this dataset, as well as verify the effectiveness of our framework for downstream tasks. QuadTrack contains $32$ video clips with $600$ frames each, a total of $19,200$ panoramic images, and $189,876$ high-quality bounding boxes with manual annotations.
The dataset contains $17$ sets of video clips as the training set, with the remaining $15$ sets as the test set. We designate $9$ sets from the test set as the new test set for evaluating generation quality and controllability, whereas the remaining $6$ sets are used as the validation set for downstream task evaluation.
Specifically, we train our generative model using the training set and generate images on the test set using the first frame's panoramic image and object bounding box prompts, then compare and evaluate the generation quality against the real images.
For controllability, we first trained OmniTrack, a multi-object tracking model for quadruped robots, using the training set. Then, we used it to track the images generated from the test set in order to evaluate the controllability of our generated outputs.
For downstream tasks, following the previous tasks TrackDiffusion~\cite{li2024trackdiffusion} and MagicDrive~\cite{gao2023magicdrive}, we add the best two video clips generated by different models into OmniTrack's training set to train the OmniTrack model, and then evaluate it on the validation set.

%% file: Contents/2_PTrack.tex
To evaluate whether the generated videos exhibit camera motion jitter consistent with real-world counterparts, 
we propose PTrack — a novel evaluation metric leveraging CoTracker3~\cite{karaev2024cotracker3} for feature point tracking to extract jitter patterns. 
CoTracker3 is a streamlined yet powerful deep learning tracking model capable of maintaining high-fidelity arbitrary point correspondence even in videos with complex motion dynamics and severe occlusion conditions.
Let the video frame resolution be \( H \times W \), and define \( G^2 \) uniformly sampled grid points \( P = \{ P_1, P_2, \dots, P_{G^2} \} \), 
where the column spacing between adjacent points is \( \Delta_x = \frac{W-1}{G} \), and the row spacing is \( \Delta_y = \frac{H-1}{G} \). 
For each sampled point, the Jectory tracked across \( T \) frames in the video is denoted as \( \mathcal{J}_{P_i} = \{ P_i^1, P_i^2, \dots, P_i^T \} \), 
where \( P_i^t = \{x_i^t, y_i^t\} \) represents the coordinates of the \( i-th \) sampled point \( P_i \) at the \( t-th \) frame.
For each tracking Jectory \( \mathcal{J}_{P_i} \), calculate its temporal variance:
\begin{equation}
\left\{
    \begin{aligned}
    \sigma_{\mathcal{J}_{P_i}} ^x = \frac{1}{T} \sum_{t=1}^{T} (x_i^t - \bar{x_i}^t )^2 \\
    \sigma_{\mathcal{J}_{P_i}} ^y = \frac{1}{T} \sum_{t=1}^{T} (y_i^t - \bar{y_i}^t )^2
\end{aligned}
\right. 
,
\end{equation}
where \( \sigma_{\mathcal{J}_{P_i}} \) is the temporal variance of the \(i-th\) sampled point \(P_i\), and \( \bar{x_i}^t \) and \( \bar{y_i}^t \) are the average coordinates of \(P_i\) in the \(x-\)direction and \(y-\)direction at the \( t-th \) frame, respectively.
The tracked trajectory \( \mathcal{J}_{P_i} \) includes both the object motion and the jitter. To evaluate the intensity of the jitter trajectory, we remove the object motion trajectory, resulting in the jitter trajectory \( \mathcal{J} \):
\begin{equation}
    \mathcal{J} = \{ \mathcal{J}_{P_i} \mid \sigma_{\mathcal{J}_{P_i}} ^x \leq Q_{0.8}(\{ \sigma_{\mathcal{J}_{P_i}}^x \}_{i=1}^{G^2})) \land \sigma_{\mathcal{J}_{P_i}} ^y \leq Q_{0.8}(\{ \sigma_{\mathcal{J}_{P_i}}^y \}_{i=1}^{G^2}) \}.
\end{equation}
Here, the expression  \( Q_{0.8}(\{ \sigma_{\mathcal{J}_{P_i}}^x \}_{i=1}^{G^2}) \) and \( Q_{0.8}(\{ \sigma_{\mathcal{J}_{P_i}}^y \}_{i=1}^{G^2}) \) represent the \( 80th \) percentile of the temporal variance of the $x$-direction and $y$-direction coordinates of all sampled points, respectively.
Finally, calculate the PTrack metric:
\begin{equation}
PTrack = \frac{1}{| \mathcal{J} |} \sum_{P_i \in \mathcal{J}} \left[ \frac{1}{T} \sum_{t=1}^{T} \left( \frac{y_{gen,p}^t - y_{real,p}^t}{\Delta y} \right)^2 \cdot \omega_p \right],
\end{equation}
where \( y_{gen,p}^t - y_{real,p}^t \) represents the difference between the generated vertical jitter and the real vertical jitter. To prevent the mean trap—where the generated video only replicates the mean trajectory of the real jitter while ignoring the fluctuation characteristics—we introduce a variance difference term \( \omega_p = (\sigma_{real,p}^y - \sigma_{gen,p}^y)^2 \) to compensate for this flaw.
In our evaluation, we set \( G = 10 \). For the $378$ 14-frame video clips in the test set, we calculate their PTrack metric and compute the average.

%% file: Contents/3_Control.tex
\textbf{Comparison with or without filtering.} 
We use a Butterworth high-pass filter to extract the high-frequency components from the object bounding box trajectories as the jitter data for the quadruped robot. We also explore the case without filtering and compare the results, as shown in Tab.~\ref{tab:filter}.
\begin{table}[h]
    \vskip-1ex
    \centering
    \renewcommand{\arraystretch}{1.2}
    \setlength{\tabcolsep}{1mm}
    \begin{tabular}{@{}l|ccccc@{}}
    \toprule
    \textbf{Method} & \textbf{PTrack ↓} & \textbf{HOTA ↑} & \textbf{MOTA ↑} &\textbf{DetA ↑} &\textbf{AssA ↑} \\
    \midrule
    No Filtering      &  17.4361  &   9.3875 &\textbf{ -19.592}	& 11.374  &7.986  \\
    Filtering       & \textbf{14.1744}   & \textbf{9.6542}&  -21.285	&\textbf{ 11.628 }&\textbf{8.2799}  \\
    \bottomrule
    \end{tabular}
    \vspace{2mm}
    \caption{\textbf{Comparison with or without filtering.} 
    ``No Filtering'' means that we directly use the object bounding box coordinates as jitter input into the framework for training and generation. }
    \label{tab:filter}
    \vskip-3ex
\end{table}
It can be observed that, except for the MOTA metric, the controllability of the model has improved in all other metrics when using the filtered data.
This is because MOTA primarily focuses on issues such as false positives, false negatives, and identity switches during object detection and tracking.
When no filtered data is used, the control signals in SOC are a fusion of object bounding box coordinates with two different encodings, which is why the MOTA metric performs better.
Additionally, we found that this method of not using filtering completely causes the model to lose the ability to control jitter.

The VJE component consists of both the filtering mechanism and the camera encoder, which serves as a crucial input to SOC. 
As the camera encoder is treated as a fixed-weight (frozen) module and integrated as a whole input, it is not feasible to ablate its sub-components separately in the current design. 
Nonetheless, as shown in Tab.~\ref{tab:vje}, we additionally conduct an ablation study using only SOC, with no filtering applied in the VJE module. 
The ablation further demonstrates the effectiveness of the filtering module in VJE.
          
\begin{table}[h]
\vspace{-0.6em}
\vskip-0.8ex
\centering
\scalebox{0.9}{ 
\setlength{\tabcolsep}{1.5mm}  
\begin{tabular}{@{} c w{c}{1cm} w{c}{1cm} w{c}{1cm} | cccc | ccc @{}}
\toprule
Setting &VJE*& SOC & PE  &  FVD ↓ & LPIPS ↓ &PSNR↑&  SSIM↑  & HOTA ↑ & MOTA ↑& PTrack↓ \\
\midrule
(a) && \checkmark & &
904.37&
0.2703&
14.33&
0.3867 &9.3886	&-20.72	&21.5676\\
(b) & \checkmark & \checkmark &  &  896.07 & 0.2620  & 14.51 & 0.3950 &9.3924 & -20.006 &15.9676 \\
\bottomrule
\end{tabular}
}
\vspace{1mm}
\caption{ VJE* refers to the configuration in which the VJE module skips filtering and performs only camera encoding on the data.} 
\label{tab:vje}
\vskip-3ex
\end{table}       

\textbf{The impact of the module on control performance.}
Through experimental verification, we found that the temporal attention blocks introduced in the SOC module are the key factor in improving the model's control effectiveness. 
Additionally, if the feature \( F_{\text{sum}} \) input to the Gated attention in the SOC module is not fused with the encoded jitter feature \( A_{\text{bg}} \), the model will lose the ability to control jitter.
This is because using only the encoding of the object bounding box positions in the gated attention causes the model to confuse the jitter of the object bounding box coordinates with the extracted jitter features.

\begin{figure}[h]
  \vskip-2ex

  \centering
  \includegraphics[width=0.99\textwidth]{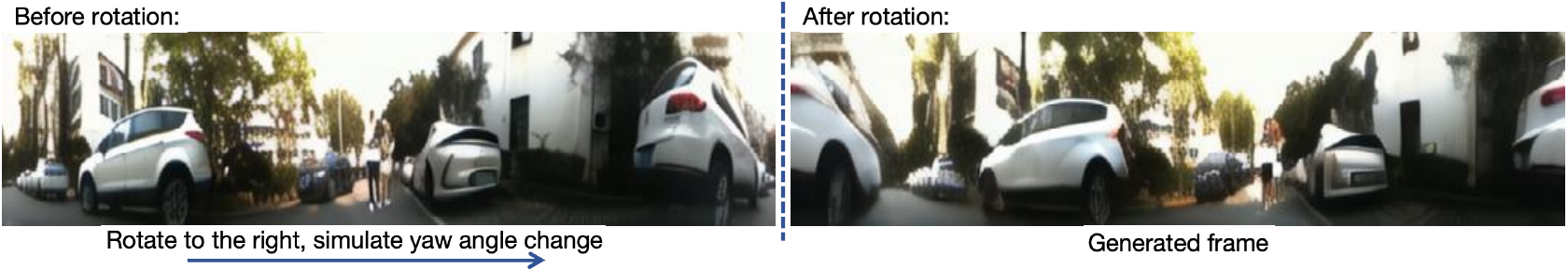}  
  \vskip-1ex
  \caption{Reasoning for simulated yaw angle changes}
  \label{fig:yaw}
  \vskip -3ex
\end{figure}

\textbf{Full 6-DoF control.} Our SOC framework supports full 6-DoF control. Internally, it uses camera encoders to represent and regulate camera pose, which has been validated in CameraCtrl to support both translation (including side-to-side sway) and rotation (yaw, pitch, and roll). 
In our current experiments, we focus on vertical jitter suppression, as it is the most prominent visual disturbance during quadruped locomotion. 
We also simulate yaw-axis variations to explore SOC’s generalization beyond vertical control.
As in Fig.~\ref{fig:yaw}, SOC remains robust under such conditions.

\textbf{Model generalization ability.} The proposed method generalizes to different types of quadruped robots. 
Our dataset includes diverse motion data from both Unitree Robotics Go2 and DEEP Robotics Lite3, with Tab.~\ref{tab:qua_ablation_supp} showing the generation performance on these two platforms. 
Lite3 exhibits more jerky motion, whereas Go2 has higher speeds but features a more limited set of movement patterns.
We will clarify the data sources of the two platforms in the final version. 
We use one model to support controllable generation for different robots.
The model is applicable to different robots exhibiting significant vertical jitter during movement.
\begin{table}[h]
\vspace{-0.8em}
  \centering
  \scalebox{0.99}{ 
  \begin{tabular}{lc|*{7}{>{\centering\arraybackslash}p{1cm}}}
    \toprule
    \textbf{Robots} &
    \textbf{Percentage} &
    FVD\,$\downarrow$ &
    LPIPS\,$\downarrow$ &
    PSNR\,$\uparrow$ &
    SSIM\,$\uparrow$ &
    HOTA\,$\uparrow$ &
    MOTA\,$\uparrow$ &
    PTrack\,$\downarrow$ \\
    \midrule
    Go2   & 44.44\% &958.87  & 0.2743 &14.1166  & 0.3691 & 7.9721 & -21.59 &16.3713  \\
    Lite3 & 55.56\% &845.15  & 0.2512 &14.8332  & 0.4152 &  11.304 &  -21.81 & 11.4282 \\
    \bottomrule
  \end{tabular}
  }
  \vspace{1mm}

\caption{ Evaluation on test data from different platforms.}
\label{tab:qua_ablation_supp}
\vskip-2ex

\end{table}

%% file: Contents/5_Visual.tex
\begin{figure*}[ht]
  \centering
  \includegraphics[width=1\textwidth]{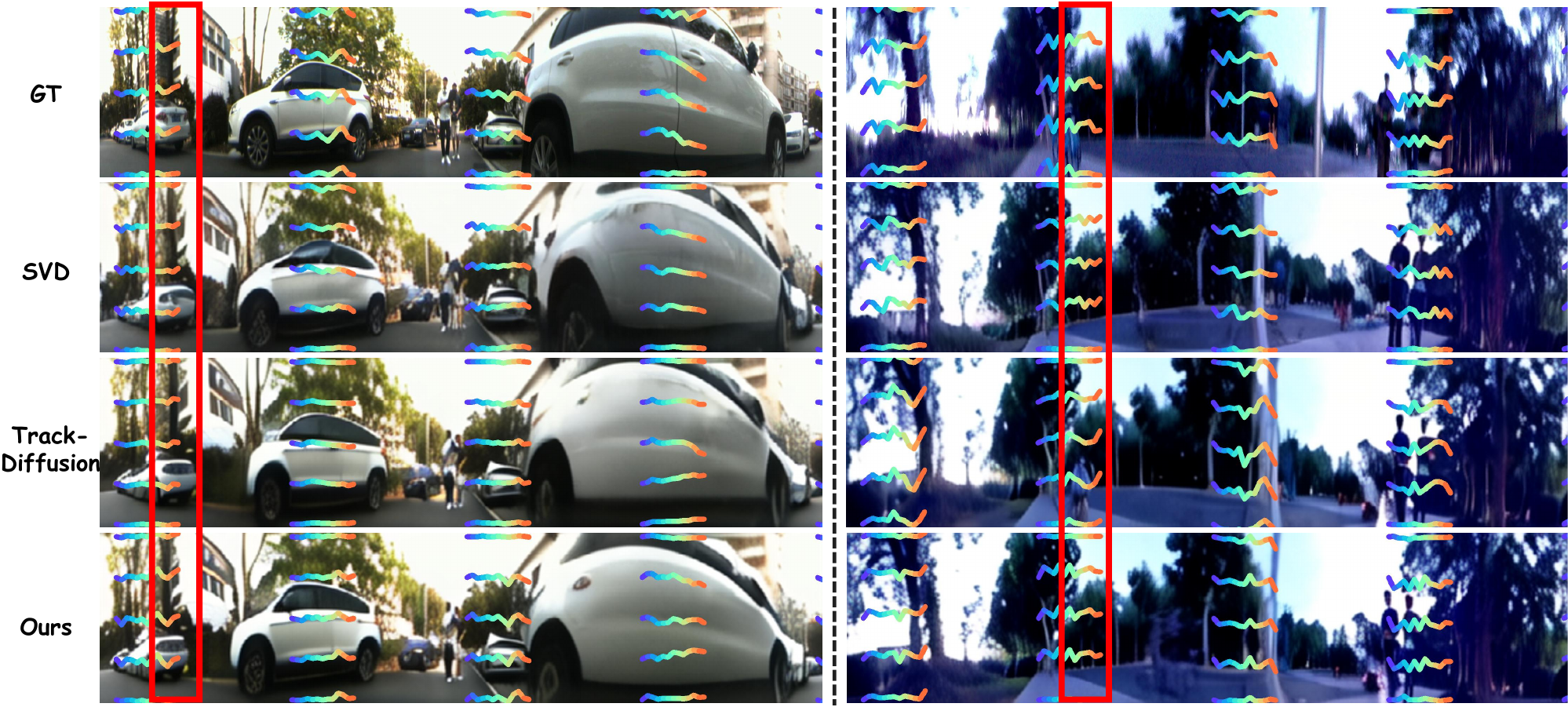}
  \vskip -1ex
  \caption{\textbf{Visual Comparison 1}}
  \label{fig:Fig1}
  \vskip -2ex
\end{figure*}
\begin{figure*}[ht]
  \centering
  \includegraphics[width=1\textwidth]{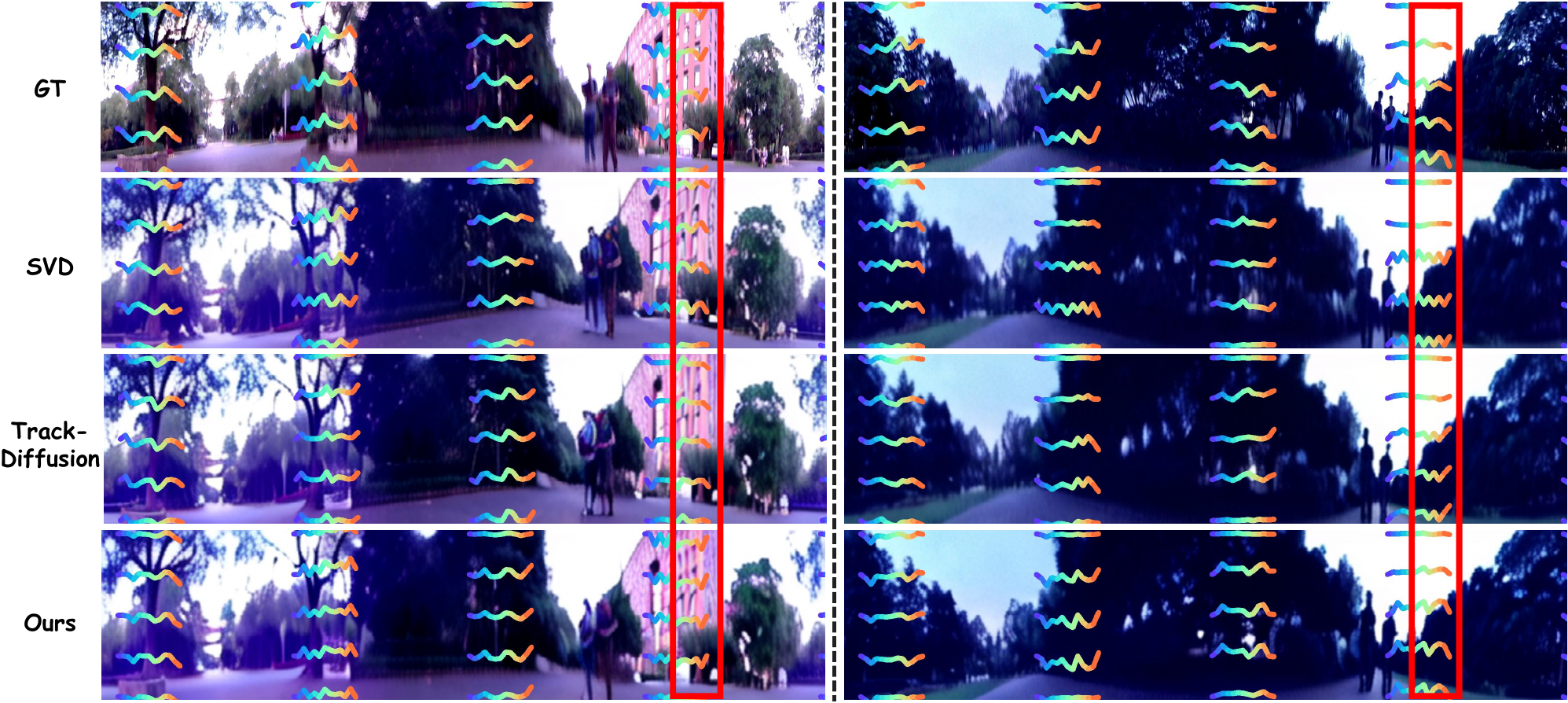}
  \vskip-1ex
  \caption{\textbf{Visual Comparison 2}}
  \label{fig:Fig2_supp}
  \vskip -2ex
\end{figure*}

\textbf{Comparison with other models.}
We present more visual results in Fig.~\ref{fig:Fig1}, \ref{fig:Fig2_supp}, and \ref{fig:Fig3_supp}, where the rainbow-colored trajectories represent the vertical jitter tracked using CoTracker3. The red boxes highlight the prominent regions. It can be observed that the video jitter generated by QuaDreamer (Ours) is closest to the ground truth video (GT).

\begin{figure*}[ht]
  \centering
  \includegraphics[width=1\textwidth]{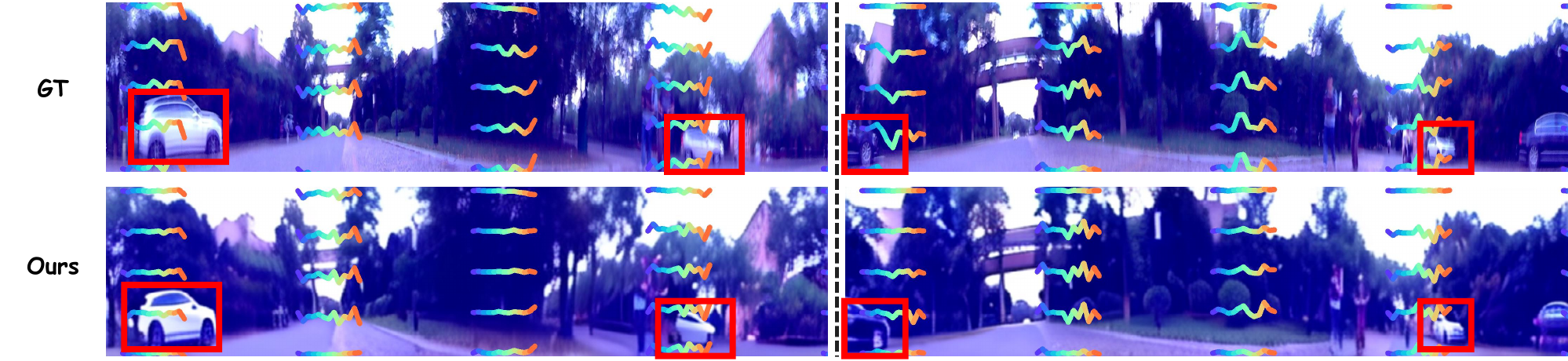}
  \vskip-1ex
  \caption{\textbf{Visualization in blurry scenes.} The object within the red box in the ground truth shows vertical blur, while the video generated by QuaDreamer effectively eliminates this issue.
}
  \label{fig:Fig4_supp}
  \vskip -2ex
\end{figure*}

\begin{figure*}[ht]
  \centering
  \includegraphics[width=1\textwidth]{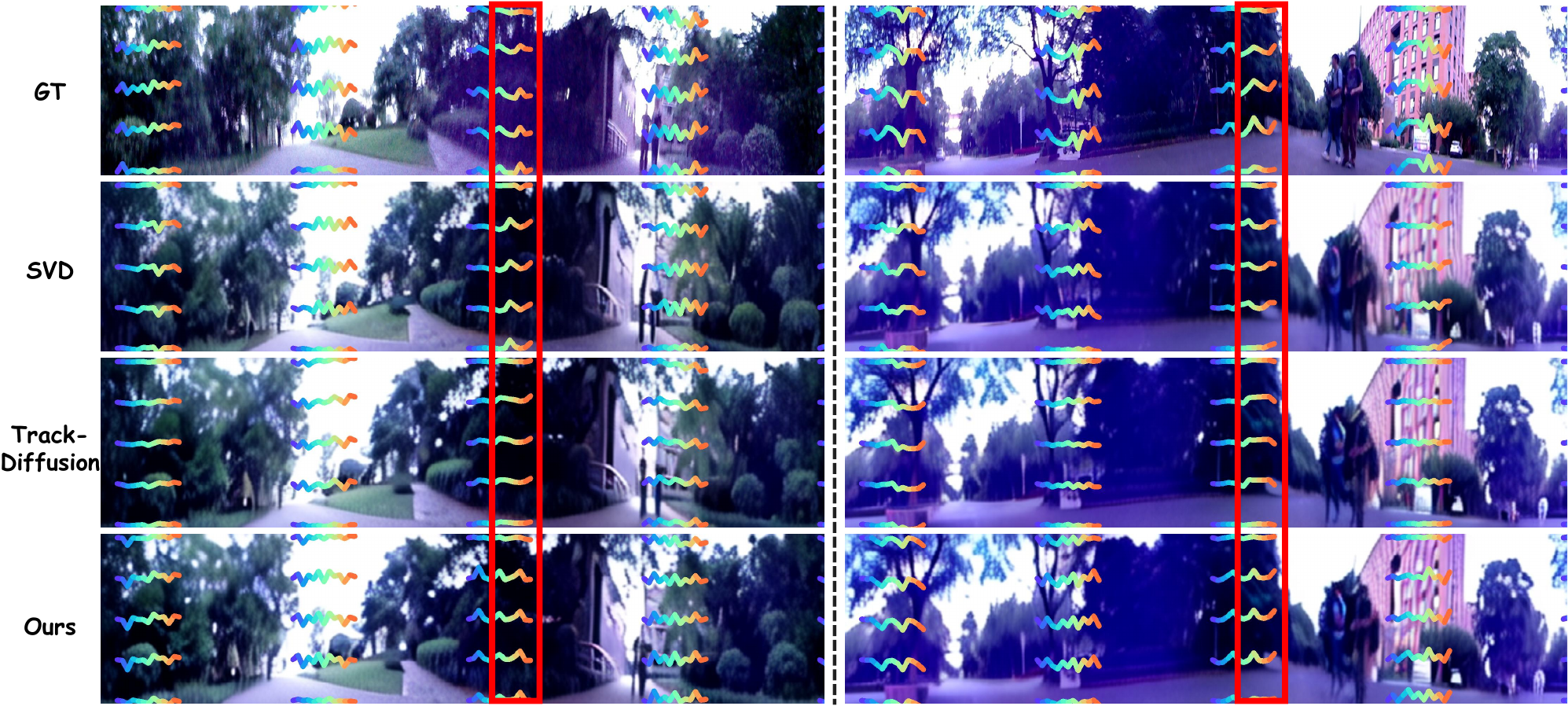}
  \vskip-1ex
  \caption{\textbf{Visual Comparison 3.}}
  \label{fig:Fig3_supp}
  \vskip -2ex
\end{figure*}

\textbf{Performance in blurry scenes.}
As shown in Fig.~\ref{fig:Fig4_supp}, due to the high-frequency vertical jitter caused by the unique gait of the robot dog, some panoramic images become blurred during capture due to long exposure times. However, the data generated by our model effectively avoids this blur, providing more effective data support for downstream tasks.

%% file: Contents/6_Significance.tex
This research has significant theoretical significance and practical value in enhancing the panoramic perception capabilities of quadruped robots. Currently, quadruped robots in complex scenarios such as inspection, rescue, and security urgently require panoramic vision systems to provide comprehensive environmental understanding. However, due to issues such as robot body motion jitter and sensor calibration difficulties, acquiring high-quality panoramic data in real-world environments is extremely costly. This data bottleneck severely restricts the training and deployment of perception models for quadruped robots. The QuaDreamer proposed in this paper, as the first panoramic data generation engine customized for quadruped robots, generates high-quality panoramic videos with vertical vibration features due to quadruped robot movement and dynamic interactions with moving targets. These videos not only provide high-quality and controllable training data for perception models but also lay the foundation for building robust panoramic vision systems. This plays a key role in advancing the autonomous interaction capabilities of quadruped robots in open environments.

From the perspective of technological innovation, the Vertical Jitter Encoding (VJE), Scene-Object Controller (SOC), and Panoramic Enhancer (PE) proposed in this study provide crucial technical support for generating panoramic videos from the perspective of a quadruped robot. VJE employs a high-pass filter to decouple the low-frequency object-relative trajectories from the high-frequency vertical jitter, and the high-frequency vertical jitter characteristics are encoded through the camera encoder. SOC utilizes a multimodal fusion mechanism with attention to control both jitter and object interactions. PE innovatively integrates frequency-domain features and spatial-domain structures through a dual-stream architecture, enhancing the quality of the generated panoramic videos. These technological innovations significantly improve the quality and control accuracy of panoramic data generation. Experiments show that synthetic data can effectively enhance the performance metrics of panoramic multi-object tracking tasks, validating its practical value in training perception models and providing an expandable technical framework for the development of future robotic simulation systems. The open-source release of this engine will further promote collaborative innovation within the quadruped robot vision community and accelerate the deployment of quadruped robot technology in complex scenarios.

%% file: Contents/7_Futurework.tex
Our work is based on Box and Jitter information; however, in real-world robotic applications, the working environment typically requires collaboration among multiple sensors (\textit{e.g.}, depth, infrared, and event cameras). Extending our framework to an RGB-X generation model is a promising direction for future exploration. 

The current work reduces the reliance on data by extracting jitter information from the coordinates of the frames. In the future, we consider directly obtaining the quadruped robot's posture information from the IMU and implementing more control effects, such as generating movements like forward/backward and turning.

Furthermore, leveraging large language models to enable generation under coarsely annotated conditions also presents a valuable and underexplored avenue.